\documentclass[letterpaper, 10 pt, journal, twoside]{ieeetran} 

\IEEEoverridecommandlockouts                              

\usepackage[utf8]{inputenc} 
\usepackage[T1]{fontenc}    
\usepackage{hyperref}       
\usepackage{url}            
\usepackage{booktabs}       
\usepackage{amsfonts}       
\usepackage{nicefrac}       
\usepackage{microtype}      
\usepackage{xcolor}         

\usepackage{subcaption,graphicx}
\usepackage{overpic}
\usepackage{wrapfig}

\usepackage{makecell}
\usepackage{array,multirow,graphicx}
\newcommand{\STAB}[1]{\begin{tabular}{@{}c@{}}#1\end{tabular}}

\usepackage{gensymb}

\usepackage{soul} 

\author{Alex Church$^{1}$, John Lloyd$^{1}$, Raia Hadsell$^{2}$ and Nathan F. Lepora$^{1}$
\thanks{Manuscript received: February, 24, 2020; Revised May, 31, 2020; Accepted June, 26, 2020.}
\thanks{This paper was recommended for publication by Editor Dan Popa upon evaluation of the Associate Editor and Reviewers' comments.}
\thanks{This work was  supported  by  an  award  from  the Leverhulme Trust on `A biomimetic forebrain for robot touch' (RL-2016-39) and an EPSRC CASE award to AC sponsored by Google DeepMind.}
\thanks{$^{1}$ AC, JL and NL are  with  the  Department  of  Engineering  Mathematics  and  Bristol  Robotics  Laboratory,  University  of  Bristol,  Bristol,  U.K.
        {\tt\small \{ac14293, jl15313, n.lepora\}@bristol.ac.uk}}%
\thanks{$^{2}$RH is with Google DeepMind.
        {\tt\small raia@google.com}}%
}

\title{\LARGE \bf
Deep Reinforcement Learning for Tactile Robotics:\\ Learning to Type on a Braille Keyboard
}

\begin{document}

\maketitle

\begin{abstract}
Artificial touch would seem well-suited for Reinforcement Learning (RL), since both paradigms rely on interaction with an environment. Here we propose a new environment and set of tasks to encourage development of tactile reinforcement learning: learning to type on a braille keyboard. Four tasks are proposed, progressing in difficulty from arrow to alphabet keys and from discrete to continuous actions. A simulated counterpart is also constructed by sampling tactile data from the physical environment. Using state-of-the-art deep RL algorithms, we show that all of these tasks can be successfully learnt in simulation, and 3 out of 4 tasks can be learned on the real robot. A lack of sample efficiency currently makes the continuous alphabet task impractical on the robot. To the best of our knowledge, this work presents the first demonstration of successfully training deep RL agents in the real world using observations that exclusively consist of tactile images. To aid future research utilising this environment, the code for this project has been released along with designs of the braille keycaps for 3D printing and a guide for recreating the experiments. A brief video summary is also available at \url{https://youtu.be/eNylCA2uE_E}.
\end{abstract}

\begin{IEEEkeywords}
Force and Tactile Sensing; Reinforecment Learning; Biomimetics 
\end{IEEEkeywords}

\section{INTRODUCTION}

\IEEEPARstart{T}{ouch} is the primary sense that humans use when interacting with their environment. Deep Reinforcement Learning (DRL) algorithms enable learning from interactions and support end-to-end learning from high-dimensional sensory input to low-level actions. However, most research has focused on vision, while a scarcity of data, sensors and problem domains relating to touch has relegated tactile research to a minor role. This trend continues even when DRL is applied to robots interacting physically with their environment, where most research uses proprioception or vision. Here we attempt to bridge the gap by positioning tactile DRL research in the context of a human-relevant task: learning to type on a braille keyboard.

The field of DRL has seen rapid progress, which in part is due to benchmark suites that allow new methods to be directly compared. These are mainly simulated environments, such as the Arcade Learning Environment \cite{Bellemare2012TheAgents}, continuous control environments \cite{Duan2016BenchmarkingControl, Tassa2018DeepMindSuite} and simulated robot-focused environments \cite{Plappert2018Multi-GoalResearch, Brockman2016OpenAIGym}. More recently, several robot benchmarking suites have been established \cite{Ahn2019ROBEL:Robots,Kumar2019OffWorldResearch,Yang2019REPLAB:Learning,Mahmood2018SettingRobot,Mahmood2018BenchmarkingRobots} with the intention of moving the field towards physically-interactive environments. 

   \begin{figure}[t]
      \centering
      \begin{overpic}[width=1.0\linewidth]{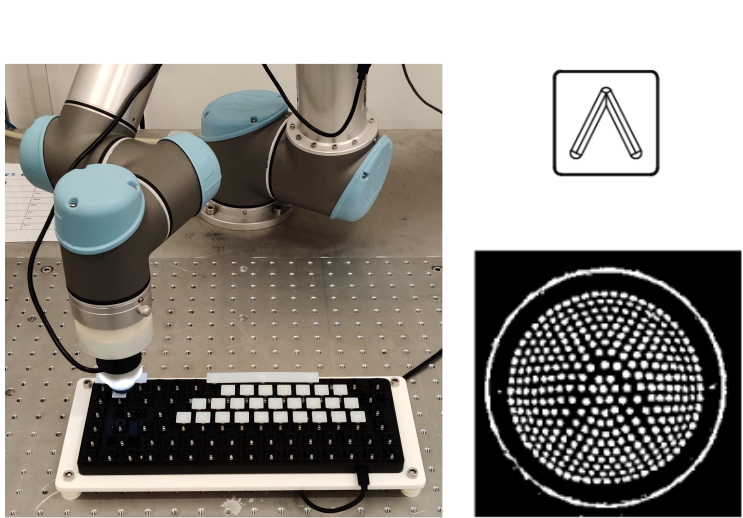}
        \put(0,65){(a)}
        \put(6,67){Robot arm with tactile sensor }
        \put(13,63){and braille keyboard}
        \put(64,65){(b) }
        \put(71,67){Example key }
        \put(72,63){(UP-arrow)}
        \put(64,40){(c)}
        \put(71,42){Tactile image}
        \put(72,38){(of UP key)}
      \end{overpic}
      \caption{Robot arm with tactile fingertip (panel a) pressing a 3D-printed UP-arrow key (panel b) resulting in a tactile image (panel c). Notice how the UP-key shape is visible in the spacing of the pins in the tactile imprint. }
      \label{fig:env_setup}
    \end{figure}
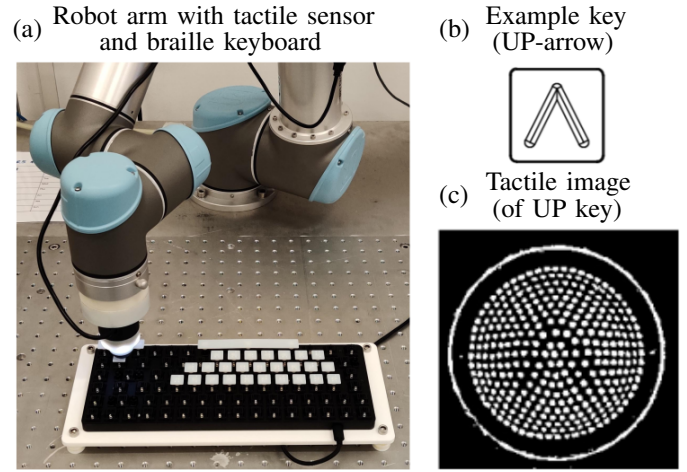

In the area of tactile robotics, there are currently no benchmarks for evaluating and comparing new methods. In this paper, we propose a challenging tactile robotics environment which we intend will serve as a tool for experimentation and fine-tuning of DRL algorithms. Furthermore, the environment challenges the capabilities of the tactile sensor in requiring sufficient sensitivity and spatial resolution \textcolor{black}{and could be used to draw comparisons between different tactile sensors}. The environment consists of a keyboard with 3D-printed braille keycaps, combined with a robotic arm equipped with a tactile sensor as its end effector. The primary task in this environment involves learning to type a key or sequence of keys, and has several useful attributes: it is goal driven, uses sparse rewards, contains both continuous and discrete action settings, is challenging in terms of exploration and requires a tactile sensor with both a high spatial resolution and high sensitivity. All of these aspects help to make the task representative of other robotics and tactile robotics challenges. The environment has also been designed to need minimal human intervention and be reasonably fast, necessary characteristics for real-world applications of DRL algorithms. \textcolor{black}{In addition, the components required for creating this environment are either widely available or 3D printed, which allows easier adoption in the tactile robotics community.}

This paper makes the following contributions. We define 4 tasks in the braille keyboard environment that progress in difficulty from arrow to alphabet keys and from discrete to continuous actions. In addition to the physical environment, we construct a simulation based on exhaustively sampling tactile data, and use this for initial validation of DRL, including agents trained with Deep Q Learning \cite{Mnih2013PlayingLearning}, Twin Delayed Deep Deterministic Policy Gradients (TD3) \cite{Fujimoto2018AddressingMethods} and Soft Actor Critic (SAC) \cite{Haarnoja2018SoftApplications}. We then demonstrate that for the majority of these tasks, DRL can be used to successfully train agents exclusively from tactile observations. However, a lack of sample efficiency for the most challenging task makes it impractical to train learning agents in the physical environment. This work thus leaves open the question of whether new or untested methods can improve this sample efficiency to a point where physical training is feasible, or whether alternative tactile sensors can represent the required tactile information in a concise enough form to allow for better sample efficiency in training. 

    \begin{figure}[t]
      \centering
      \includegraphics[width=1.0\linewidth]{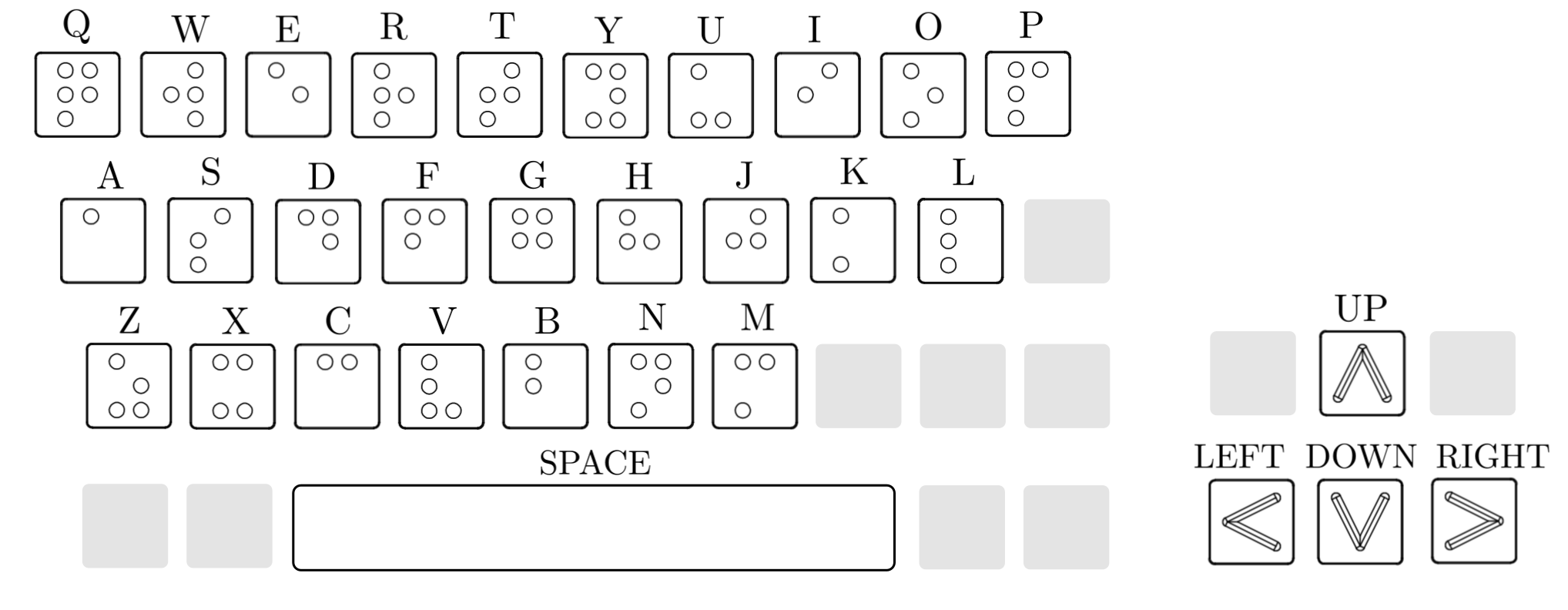}
      \caption{Braille alphabet used for the tactile keyboard environment. The space bar is a blank character, which will cause deformation of the tactile sensor. Grey squares indicate positions where no deformation of the sensor will occur. A simpler task can be defined using only the 4 arrow keys. }
      \label{fig:braille_keyboard}
    \end{figure}

\section{RELATED WORK}

A common approach for applying DRL to robotics is to train on accurate simulations of the robot, then transfer those learned policies to the real world, bridging what is known as the `reality gap' \cite{James2018Sim-to-RealNetworks, OpenAI2018LearningManipulation}. However, a serious issue for this approach is that the physical properties of artificial tactile sensors are highly challenging to simulate, as evidenced by a lack of use in the field compared to {\em e.g.} simulated visual environments. This issue is compounded as the task complexity increases, particularly with dynamic environments. This necessitates solutions that are capable of being trained directly on a real robot to make progress on DRL for tactile sensing applied to physical interaction.

DRL has been applied to object manipulation, particularly with dexterous robotic hands. However, the most successful examples of learning in this area do not utilise tactile data \cite{Rajeswaran2018LearningDemonstrations, OpenAI2018LearningManipulation, OpenAI2019SolvingHand}, but instead deploy simulations to accelerate learning. In these cases, observations are often made up of joint angles and object positions/orientations; in practise, obtaining this information from real-world scenarios has required complicated visual systems. In their work, OpenAI state that they specifically avoid using sensors built into the physical hand, such as tactile sensors, as they are subject to ``state-dependent noise that would have been difficult to model in the simulator'' \cite{OpenAI2018LearningManipulation}.

Most prior research applying RL to tactile sensing has used the SynTouch Biotac, a taxel-based tactile sensor the size of a human fingertip. Van Hoof et al. \cite{vanHoof2016StableData} demonstrated that tactile data can be used as direct input to a policy network, trained through RL, for a manipulation task on a physical robot.  In \cite{Chebotar2016Self-supervisedLearning} dimensional reduction is used to simplify the tactile input, resulting in successful regrasping of an object utilising tactile data. In \cite{Huang2019LearningLearning}, tactile and proprioceptive data are combined to achieve the task of gentle object manipulation, where exploration is improved by using tactile data to form both a penalty signal and as a target for intrinsic motivation.  

Recent work has combined tactile image observations from the Gelsight optical tactile sensor with proprioceptive information to use DRL to learn surface following \cite{Lu2019SurfaceSensor}. Optical tactile sensors provide tactile information in a form that is well-matched to modern deep learning techniques that have been honed on computer vision challenges. Given the recent successes of optical tactile sensors with deep learning \cite{Lepora2018FromSensor, Yuan2017Shape-independentSensor, Calandra2017TheOutcomes, Calandra2018MoreTouch, Hogan2018TactileTransformations, Lepora2020IEEERAM}, the combination of deep reinforcement learning and optical tactile sensing appears a promising avenue for future research.


    \begin{figure}[t!]
      \centering
      \includegraphics[width=1.0\linewidth]{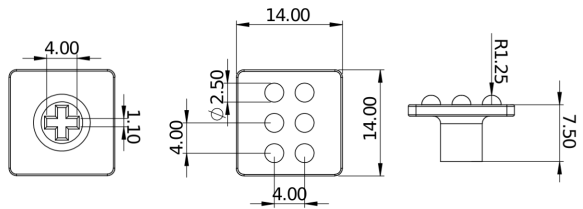}
      \caption{Dimensions of the braille keycaps in mm, designed for cherry MX mechanical switches.}
      \label{fig:keycap_design}
    \end{figure}

\section{HARDWARE}

    \begin{figure*}[t!]
      \begin{overpic}[width=1.0\linewidth]{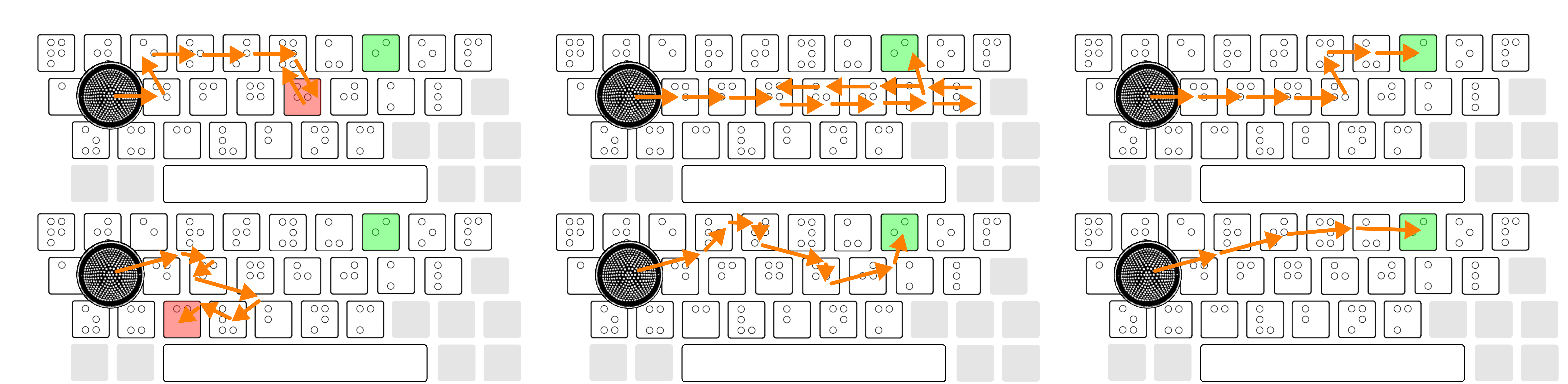}
        \put(8,23){(a) Start of Training}
        \put(40,23){(b) Middle of Training}
        \put(76,23){(c) End of Training}
        \put(0,14){\rotatebox{90}{Discrete}}
        \put(0,1){\rotatebox{90}{Continuous}}
      \end{overpic}
      \caption{Visualisation of both the discrete action (top) and continuous action (bottom) tasks. The sensor is initialised in a random position with the task of pressing a randomly initialised goal key (green). Orange arrows indicate the actions taken at each step. Initially, the policy networks cause random actions that results in an incorrect key press (red). After some training, actions that lead to a correct key press are found but may be sub-optimal, often with cyclic movements. Eventually, policies should converge to a near-optimal path between the initial position and goal key. For the discrete case, actions are given by a DQN agent trained for 0, 30 and 100 epochs. The continuous case is an illustrative depiction of successful training.}
      \label{fig:training_graphic}
    \end{figure*}
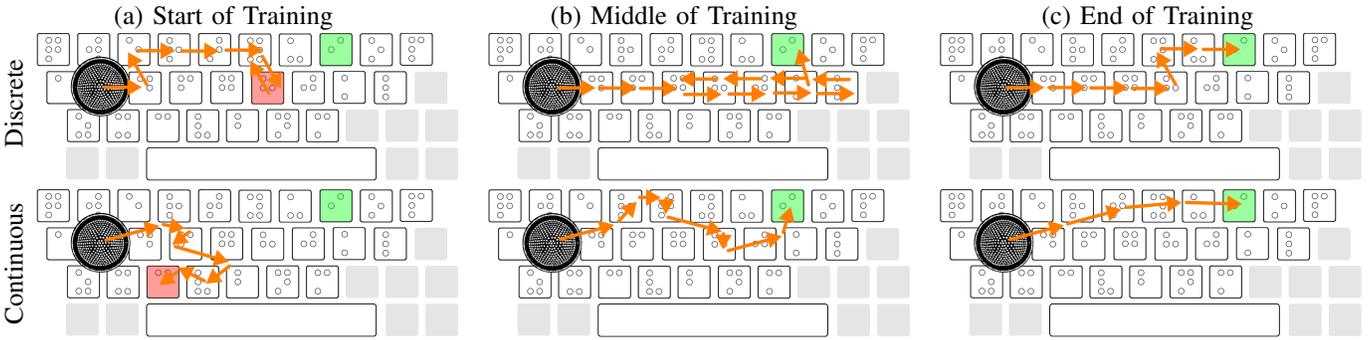
    
\subsection{Custom biomimetic tactile fingertip}

The BRL tactile fingertip (TacTip) \cite{Ward-Cherrier2018TheMorphologies} is a low-cost, robust, 3D-printed optical tactile sensor based on a human fingertip. As the current task involves interpreting braille keycaps, we require a tactile sensor with high spatial resolution that makes contact with an area that can lie inside a standard keycap. Conventionally, the TacTip has 127 pins on a 40\,mm-diameter tactile dome~\cite{Ward-Cherrier2018TheMorphologies}, which is too large for this task. The sensor tip was thus redesigned for this work to fit 331 pins of radius $0.625\,$mm onto a $25\,$mm-diameter dome of depth $7.2\,$mm. The pins are arranged such that there is a sub-mm space between them. An example tactile image obtained from this modified sensor is shown in Figure \ref{fig:env_setup}c, where adaptive thresholding is used to process the raw tactile image into a binary image that makes the deformation more apparent and mitigates any changes of lighting inside the sensor.



Following recent work with this tactile sensor~\cite{Lepora2018FromSensor}, the pre-processed image captured by the sensor is directly passed into the machine learning algorithms. This removes the need for pin detection and tracking algorithms that were necessary in previous work, and enables the advantages of convolutional neural networks to be applied to tactile data. 

\subsection{Braille Keyboard}

For the tasks considered here, we chose a good-quality keyboard with a fairly stiff key switch: the DREVO Excalibur 84 Key Mechanical Keyboard with Cherry MX Black switches. These switches have a linear pressing profile and require $0.6\,$N of force and $2\,$mm of travel before actuation occurs. Furthermore, Cherry keys have good build quality which improves consistency across keys. The dimensions of our 3D-printed keycaps are shown in Figure \ref{fig:keycap_design} and the full set of keys used throughout this work are shown in Figure \ref{fig:braille_keyboard}. The keyboard was chosen in order to allow the tactile robot to make exploratory contacts with a key before pressing it.

\subsection{Robotic Arm}
For these experiments we use a Universal Robotics UR5 robotic arm and its in-built inverse kinematics to allow for Cartesian control. This makes the environment relatively agnostic to the type of robotic arm used, with variation only arising subject to error accumulating from the lack of precision in the inverse kinematics controller. Other than the robotic arm, the environment consists of components that are either 3D printed or widely available.

\section{BENCHMARK TASKS}

Within this tactile keyboard environment, we propose 4 tasks of progressive difficulty that have the same underlying principles:
\begin{itemize}
    \item Discrete actions: Arrow keys.
    \item Discrete actions: Alphabet keys.
    \item Continuous actions: Arrow keys.
    \item Continuous actions: Alphabet keys.
\end{itemize}
The distinction between arrow and alphabet keys is shown in Figure \ref{fig:braille_keyboard}. Note that the space key is included in alphabet tasks and is represented by a blank character. \textcolor{black}{ Continuous actions constitute a positional change in the $x,y$ axis and a tap depth in the $z$ axis, where the sensor performs a downward movement and returns to a predefined height. This is performed sequentially to reduce the potential for damage occurring.} An illustration of successful training of a RL agent is depicted in Figure~\ref{fig:training_graphic} over the alphabet keys for both discrete and continuous actions.

The proposed task for this environment is to successfully press a goal key, which is randomly selected each episode from a subset of keys on a braille keyboard, and to do so without actuating any other keys. The start location is randomised for each episode. Hence, the agent must first determine its location, then navigate to the target position and finally press the target key. Positions of all the keys are learnt implicitly within the weights of a neural network. A positive reward of $+1$ is given for successfully pressing a goal key; an incorrect key press results in episode termination and a reward of~$0$. Successful learning of this task will result in an average episodic return close to $1$, dependent on the exploration parameters of the RL agent. 

The arrow keys offer a natural way to define an easier task, since they are located away from the other keys, are fewer, and span a smaller area, which greatly reduces the size of the state space. An agent should therefore encounter more positive rewards during random exploration. Moreover, the tactile features on the arrow keys are more distinct, easing their interpretation from tactile data.

The alphabet (plus space) keys give a far more challenging task. As the state space is relatively large, learning over the full set of alphabet keys requires an approach such as Hindsight Experience Replay (HER) \cite{Andrychowicz2017HindsightReplay} to improve data efficiency. Also, depending on the position of the sensor, some keys can be indistinguishable from each other, giving subtleties from perceptual aliasing. This feature also makes it harder to learn continuous action policies, since when using discrete actions the sensor can be arranged to be always located above the centre of a key. 

\begin{table}[htbp]
	\centering
	\caption{\label{tab:hyperparams} DRL and network hyperparameters.}
	
	\begin{tabular}{|c|l|c|c|c|}
		\hline
		& \multicolumn{1}{l|}{ \textbf{Param} } &  \multicolumn{3}{c|}{ \textbf{Shared} }   \\ 
		\hline 
		
		\multirow{9}{*}{\STAB{\rotatebox[origin=c]{90}{ \textbf{Network} }}}
		& Input dim        &  \multicolumn{3}{c|}{ [100, 100, 1] }        \\
		& Conv filters     &  \multicolumn{3}{c|}{ [32, 64, 64] }         \\
		& Kernel widths    &  \multicolumn{3}{c|}{ [8, 4, 3]    }         \\
		& Strides          &  \multicolumn{3}{c|}{ [4, 2, 1]    }         \\
		& Pooling          &  \multicolumn{3}{c|}{ None         }         \\
		& Dense layers     &  \multicolumn{3}{c|}{ [512,]       }         \\
		& Output dim       &  \multicolumn{3}{c|}{ num actions   }         \\
		& Activation       &  \multicolumn{3}{c|}{ \textit{ReLU} }        \\ 
		& Initialiser      &  \multicolumn{3}{c|}{ variance scaling (scale=1.0) }        \\ 
		\hline 
		
		\multirow{8}{*}{\STAB{\rotatebox[origin=c]{90}{ \textbf{Control} }}}
		& Epoch steps          &  \multicolumn{3}{c|}{ 250    }     \\
		& Replay size          &  \multicolumn{3}{c|}{ $10^5$ }     \\
		& Update freq          &  \multicolumn{3}{c|}{ 1      }     \\
		& n/ Updates           &  \multicolumn{3}{c|}{ 2      }     \\
		& Batch size           &  \multicolumn{3}{c|}{ 32     }     \\
		& Start steps          &  \multicolumn{3}{c|}{ 200    }     \\
		& Max ep len           &  \multicolumn{3}{c|}{ 25     }     \\ 
		& Optimizer            &  \multicolumn{3}{c|}{ Adam   }     \\
		\hline \hline 
		
		&  &  \textbf{DQN}  &  \textbf{SAC}   &  \textbf{TD3} \\
		\hline 
		
		\multirow{4}{*}{\STAB{\rotatebox[origin=c]{90}{ \textbf{RL} }}}
		& Discount ($\gamma$)    & 0.95         &  \makecell{0.6 (disc) \\ 0.95 (cont)}   & 0.95    \\
		& Polyak  ($\tau$)       & 0.005        & 0.995              & 0.995   \\
		& LR ($\eta$)      & $5 \times 10^{-4}$ & $5 \times 10^{-4}$ & $5 \times 10^{-5}$ \\
		& Alpha LR ($\eta_{\alpha}$)  & n/a     & $1 \times 10^{-3}$ & n/a \\ \hline 
		
	    \multirow{4}{*}{\STAB{\rotatebox[origin=c]{90}{ \textbf{Explore} }}}
		& Initial $\epsilon$      & 1.0  & n/a & 1.0  \\
		& Final $\epsilon$        & 0.1  & n/a & 0.05 \\
		& $\epsilon$ Decay steps  & 2000 & n/a  & n/a \\
		& Target entropy          & n/a  & \makecell{0.139 (disc) \\ -6 (cont)} & n/a \\
		\hline
	\end{tabular}
\end{table}

\section{INITIAL ASSESSMENT USING SUPERVISED DEEP LEARNING}

For a first step, we check that the braille keys can be interpreted by the tactile sensor without exerting enough force to activate the button. To test this, we used supervised learning to classify all keys on the braille keyboard (shown in Figure \ref{fig:braille_keyboard}). Overall, we used 100 samples per key for training and 50 samples per key for validation, resulting in 3100 training images and 1550 validation images. 

To make the task more representative of how humans perform key presses, we introduced small random perturbations in the action. Each tactile image was collected at the bottom of a tapping motion, where the sensor was positioned $3.5+\Delta z\,$mm above the centre of each key, then moved downwards $5\,$mm, finishing above the $2\,$mm activation point for a key press. A random variation $\Delta z$ sampled from the interval $(-1,0)$\,mm was used to represent uncertainty in the key height, which ranges from `barely touching' to `nearly activating' the button. A similar random horizontal $(x,y)$-perturbation was sampled from ranges $(-1,+1)\,$mm and a random orientation perturbation was sampled from ($-10\degree,10\degree$) to add further variation in the collected data. 

Tactile images were cropped and resized down to $100\times100$ pixels (from $640\times480$ pixels), then adaptive thresholding was used to binarize the image; in addition to reducing image storage, this also helps accentuate tactile features and mitigate any issues from changes in internal lighting. 

The network used in this task follows the architecture used for learning to play Atari games, originally proposed in \cite{Mnih2015Human-levelLearning}. The same network architecture is used for all reinforcement learning tasks (details given in Table \ref{tab:hyperparams}).  For supervised learning, we perform data augmentation including random shifting and zooming of the images. We also use early stopping, a decaying learning rate, batch normalization on the convolutional layers after the activation and dropout on the fully connected layers. 

A near perfect overall accuracy of $99\%$ was achieved, demonstrating that the braille can be interpreted by the tactile sensor without activating the button, even when there is significant uncertainty about how the key is pressed.

\section{SIMULATED ENVIRONMENT}

Even though our focus in this work is DRL on a physical tactile robot, it is useful to have a simulated task that is similar to the physical environment yet runs much faster. While the policies trained in simulation may not be directly transferable to the physical environment, the parameters found should guide successful training of policies on the physical robot.

In the simulation of the discrete task, actions lead to locations on a virtual keyboard with a known label. The label is then used to retrieve a tactile image of the same label from the data collected for the initial assessment of supervised learning (Section VI). Thus the simulation accurately matches the physical environment, where the label of each key may not be known but a tactile image observation can be collected that will resemble the one obtained from the simulated environment. In practice, this simulated discrete environment is more complicated than the physically interactive environment, because of the perturbations introduced in the data collection that we choose not to introduce during reinforcement learning on the physical robot.

The simulation of the continuous task is more difficult to represent, because we do not have labels for every location on the keyboard. To approximate the physical environment, we collect tactile observations over a dense grid of locations spanning the keyboard, with those location stored (using $3\,$mm intervals over the $x,y$-dimensions and $1\,$mm intervals over $z$). Along with the tactile image observations,  we also record whether a key has been activated or not. During simulation, the position of the virtual sensor is known, which allows for a tactile image to be sampled from the collected data with the minimum Euclidean distance from the virtual sensor position. 

In both cases, the simulated environment ignores effects such as error that accumulates over long sequences of robot arm moves, sensor drift, changing lighting conditions, movement of the keyboard and any other information that is not represented during the data collection stage. However, whilst these effects can occur on a real robot, the simulated tasks still offer useful information for hyper-parameter tuning and initial validation of RL algorithms.

    \begin{figure*}[t!]
      \centering
      \includegraphics[width=1.0\linewidth]{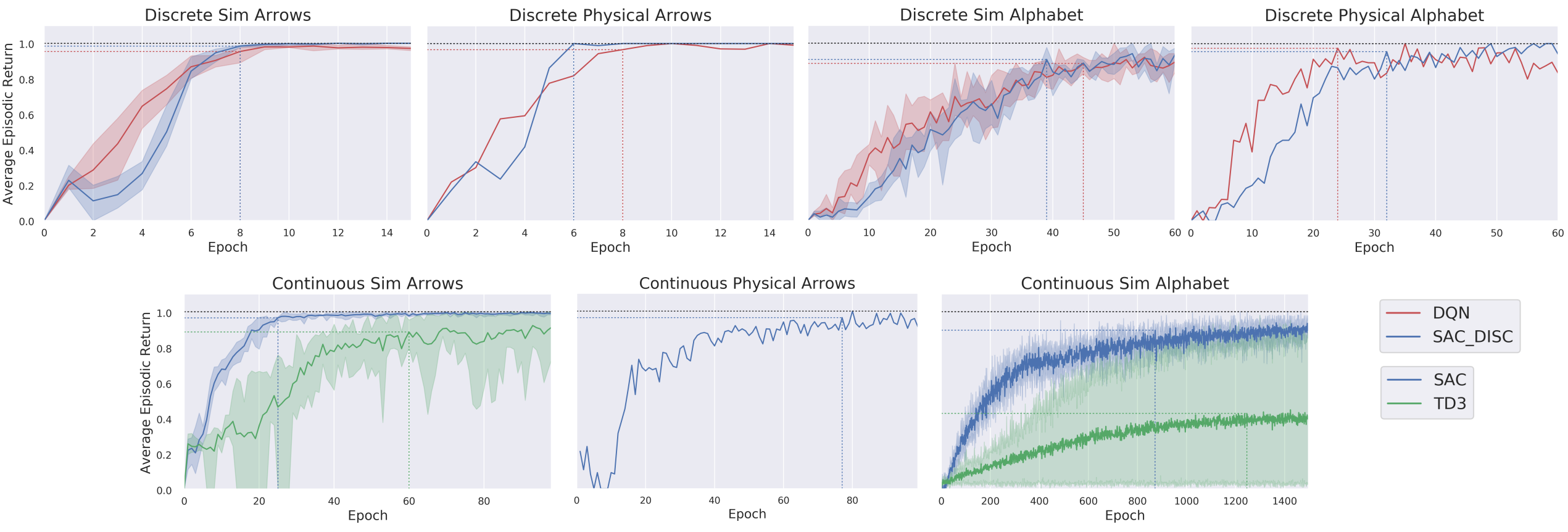}
      \caption{Training curves showing the average episodic return for all tasks, discrete task results are shown on the top row, continuous task results are shown on the bottom row. Each epoch corresponds to 250 steps in the environment. For the simulated tasks results are averaged over 5 seeds, the physical tasks show results for 1 seed. \textcolor{black}{Coloured dashed lines represent the first point at which episodic return reaches 95\% of the maximum reached throughout training.}}
      \label{fig:results}
    \end{figure*}

\section{TACTILE REINFORCEMENT LEARNING}

\subsection{Reinforcement Learning Formulation}

To define the problem, we represent the proposed tasks in a standard Markov Decision Process (MDP) formulation for reinforcement learning. An agent interacts with an environment at discrete time steps $t$, consisting of a state $s$, action $a$, resulting reward $r$, resulting state $s^\prime$ and terminal signal $d$. Actions $a$ are sampled from policy distribution $\pi(a|s)$ represented by a neural network. 

States $s$ are represented as a combination of a tactile image observation $o$ from the sensor, goal $g$ and previous action $a_{t-1}$. Goals $g$ are selected randomly per episode and represented as a one-hot array, where the `hot' value represents the class label of a target key. The previous action $a_{t-1}$ is required to avoid an agent becoming stuck in cyclic patterns when no tactile deformation is present on the sensor. Both are concatenated into the policy networks after any convolutional layers. Reward $r$ is sparse and binary, with $r=1$ when a correct button is actuated, otherwise $r=0$. Activation of any button, correct or incorrect, results in a terminal signal $d=1$ that resets the environment with the sensor moving to a new random location on the keyboard.

In the discrete tasks, actions are selected from the set $\mathcal{A} = \{\texttt{UP, DOWN, LEFT, RIGHT, PRESS}\}$. \textcolor{black}{For each movement action the tactile sensor is re-positioned above the centre of a neighbouring key where a tap action (which does not activate a key) is performed and a tactile image for the next state $s^\prime$ is gathered. The $\texttt{PRESS}$ action moves the sensor in the $z$ axis by $-8\,$mm to activate a key.} 

In the continuous tasks, each action is selected from $\mathcal{A} = \{\Delta{x}, \ \Delta{y}, \ \Delta{z} \}$ where $\Delta{z}$ corresponds to a tapping depth. For practical reasons, actions are bound to a finite range dependent on task. For the arrow task, the enforced $\Delta{x}$ and $\Delta{y}$ bounds are between $\pm 10\,$mm and for the alphabet task between $\pm20\,$mm. In both the alphabet and arrow tasks, $\Delta{z}$ is bound to the range $[2,8]\,$mm to ensure safe key actuation.

\subsection{Reinforcement Learning Algorithms}

There are several popular DRL algorithms in common use \cite{Mnih2013PlayingLearning, Schulman2015TrustOptimization, Schulman2017ProximalAlgorithms, Lillicrap2015ContinuousLearning, Fujimoto2018AddressingMethods, Haarnoja2018SoftActor, Abdolmaleki2018MaximumOptimisation}. However, there are two major problems holding back the application of DRL to physical robotics: poor sample efficiency and brittleness with respect to hyper-parameters. Generally, on-policy methods such as Trust Region Policy Optimisation (TRPO) and Proximal Policy Optimisation (PPO) sacrifice sample efficiency to gain stability and robustness to hyper-parameters, making them difficult to apply to physical robots. Since its introduction, Deep Q-Learning (DQN) has had various improvements to address these issues, some of which are amalgamated in RAINBOW \cite{Hessel2017Rainbow:Learning}. Off-policy entropy-regularised reinforcement learning has also separately addressed these issues with Soft Actor Critic (SAC) \cite{Haarnoja2018SoftActor} and Maximum a Posterior Policy Optimisation (MPO) \cite{Abdolmaleki2018MaximumOptimisation}. Both of these offer good sample efficiency, robustness to hyper-parameter selection and work with either continuous or discrete action spaces. SAC has led to the most follow-up research and a modified version has demonstrated successful training when applied to physical robots \cite{Haarnoja2018SoftApplications}. Twin Delayed Deep Deterministic Policy Gradients (TD3) \cite{Fujimoto2018AddressingMethods} was developed concurrently to SAC and offers similar performance. 

For discrete-action tasks, we use DQN with the double \cite{Hasselt2016DeepQ-Learning} and dueling \cite{Wang2015DuelingLearning} improvements, and SAC adapted for discrete actions. For continuous-action tasks, we use TD3 and SAC; since SAC is applicable to both types of action space, some hyper-parameters are transferable over all tasks.

Another barrier to the application of DRL to robotics is when environments require challenging exploration combined with sparse rewards, as convergence is then very slow. A common technique commonly used to alleviate this is to use dense and shaped rewards \cite{Popov2017Data-efficientManipulation, Mahmood2018BenchmarkingRobots, Yang2019REPLAB:Learning, Haarnoja2018SoftApplications}. However, this can require domain-specific knowledge and bias learning algorithms towards sub-optimal policies. Hindsight Experience Replay (HER) helps address this problem by replaying episodes stored in a buffer while varying the goal from what was initially intended. Here we find HER significantly improves performance and sample efficiency for all considered tasks.

Several other adjustments were also made to optimise performance. For all algorithms except TD3, the ratio of optimisation steps per environment step was increased to $2:1$ to improve efficiency because of the greater cost of environment versus optimisation steps; however, this hindered the stability of TD3. Additionally for both DQN and TD3 we linearly decayed the exploration parameter $\epsilon$ for an initial number of exploration steps. For SAC and TD3, polyak averaging was used for target networks, with a coefficient of $\tau=0.995$, although we did find that SAC could also learn well with hard target updates.
    
\section{EVALUATION METRICS}

\textcolor{black}{To quantify the performance when training an agent on this benchmark, we introduce several evaluation metrics particular to this task. These can be used to compare results of this benchmark when using alternative algorithms or alternative sensors. The evaluation metrics introduced are:}

\textcolor{black}{
\begin{itemize}
    \item \textbf{Convergence Epoch:} Measures the first epoch in which 95\% of the maximum performance  across the full training run is achieved and serves as an indication of sample efficiency during training. This is subject to some noise and works best when results are averaged over multiple seeds or when test results are averaged over a relatively high number of episodes ($\textgreater10$).
    \item \textbf{Typing Accuracy:} Measures the accuracy when typing example sentences or sequences of keys. As a reward of 1 is given for correct key presses this metric is directly correlated with the average return.
    \item \textbf{Typing Efficiency:} Measures the number of steps taken to press a key and gives an indication as to whether the policies learnt are near optimal.
\end{itemize}
}

When measuring typing accuracy and efficiency for the arrow task all 24 permutations of \{\texttt{UP, DOWN, LEFT, RIGHT}\} are evaluated. For the alphabet task, we provide 10 randomly generated permutations of the alphabet. This tests that a mapping between all keys is stored implicitly within the weights of the policy neural networks. During evaluation, the sensor is initialised in a random starting position for each sequence and the sensor position is not randomly reset after episode termination when a single key is pressed unlike during training. \textcolor{black}{Furthermore, during evaluation we used deterministic actions to avoid mispressed keys from unnecessary exploration. This procedure gives a more consistent value in comparison with using average episodic return, which allows better comparisons to be drawn.}

\section{RESULTS}

\subsection{Results for Discrete Tasks}

For the discrete arrow key and alphabet key tasks, we find that both DQN and Discrete SAC rapidly converge to accurate performance in simulation and on the physical robot. Testing over multiple seeds in simulation shows that this learning is stable and consistently converges to an average episodic reward of near 1. The training curves for the discrete tasks for both the arrow and alphabet key environments (Figure \ref{fig:results}) show that the task can be learned in all cases.

For the arrow task, asymptotic performance is achieved within 15 epochs, which is approximately equal to 1 hour of training time on the physical robot. For the alphabet task, convergence is longer, with asymptotic performance achieved within 60 epochs, or approximately 4 hours of training time on the physical robot. Similar sample efficiency and final performance is found for both discrete SAC and DQN across all trained agents. Whilst final performance appears slightly higher for discrete SAC, this can be explained by the target entropy causing lower levels of exploration when the agent nears convergence. Decaying the exploration parameter ($\epsilon$) to a lower value during training for the DQN agents, or evaluating with deterministic actions, results in similar final performance to discrete SAC.

\subsection{Results for Continuous Tasks}
    
The continuous control tasks are far more challenging due to their larger state spaces. For the arrow task in simulation, we find SAC is still able to achieve an asymptotic performance of near $1.0$ within 50 epochs. TD3 is also able to achieve performance of near $1.0$ within 100 epochs; however, training is less stable and is not robust to hyper-parameter changes. Due to this instability we do not evaluate TD3 in the physical environment. The training curves are shown in Figure \ref{fig:results} (bottom panels) for all continuous tasks that we were able to run to completion.  

For the continuous arrow task, continuous control is not as well represented in simulation, this is shown by the left and middle panels in Figure \ref{fig:results} (bottom) not being accurate matches. We find convergence to a slightly lower average episodic return of $0.95$ after about 76 epochs compared to 25 epochs in simulation. 

The alphabet task again offers an increase in task complexity. For a single-seeded run in simulation, we achieved a final average episodic return of $\sim0.9$ for the continuous SAC algorithm and $\sim0.8$ for TD3. This final performance takes significantly longer to achieve than all other tasks, with convergence occurring around 872 epochs for SAC and 1246 epochs for TD3. This will correspond to approximately 60 hours of physical robotic training time and is currently not feasible given laboratory operating constraints.
    
\subsection{Performance on Evaluation Metrics}

\begin{table}[t!]
	\centering
	\caption{\label{tab:eval} Results for trained agent evaluation.}
	\setlength\tabcolsep{4pt}
	\begin{tabular}{|c|c|c|c|c|c|}
		\hline 

		& \textbf{Task}  &  \textbf{Algorithm} &  \textbf{Steps} &  \textbf{Accuracy} & \makecell{\textbf{Convergence} \\ \textbf{Epoch}}   \\ 
		\hline \hline
		
		\multirow{8}{*}{\STAB{\rotatebox[origin=c]{90}{ \textbf{Simulation} }}}
		& \multirow{2}{*}{Disc Arrow} &   DQN        & 230 & 1.0 &  8  \\
		&                             &   SAC\_DISC  & 234 & 1.0 &  8  \\
		\cline{2-6}		
		& \multirow{2}{*}{Disc Alpha} &   DQN        & 1722 & 0.981 & 45  \\
		&                             &   SAC\_DISC  & 1649 & 0.940 & 39  \\
		\cline{2-6}
		& \multirow{2}{*}{Cont Arrow} &   TD3        & 140  & 0.906 & 60  \\
		&                             &   SAC        & 133  & 1.0   & 25    \\
		\cline{2-6}
		& \multirow{2}{*}{Cont Alpha} &   TD3        & 1169 & 0.811 &  1246  \\
		&                             &   SAC        & 1193 & 0.933 &  872   \\
		\hline \hline

		\multirow{5}{*}{\STAB{\rotatebox[origin=c]{90}{ \textbf{Physical} }}}
		& \multirow{2}{*}{Disc Arrow}      &   DQN        & 246 & 1.0 &  8  \\
		&                                  &   SAC\_DISC  & 246 & 1.0 &  6  \\
		\cline{2-6}		
		& \multirow{2}{*}{Disc Alpha}    &   DQN        & 1722 & 0.992 & 24  \\
		&                                &   SAC\_DISC  & 1649 & 0.985 & 32 \\
		\cline{2-6}
		& Cont Arrows  	&   SAC       &  364  & 0.938 & 76 \\
		\hline

	\end{tabular}
\end{table}

\textcolor{black}{An overview of the results across all algorithms used within this work are given in Table \ref{tab:eval}. These can be used as a comparative point for future work using this benchmark.} For discrete tasks, we find that high typing accuracy is possible across both simulated and physical environments. In evaluation, both SAC\_DISC and DQN have comparable results for typing accuracy and typing efficiency. In simulation, continuous tasks display a large reduction in the number of steps taken to activate a goal key due to the more efficient path across the keyboard that the sensor can take. However, this comes at the cost of a reduction in accuracy. These more-efficient policies are not observed in the physical task, likely due to more variety in tactile observations causing less certainty in the selected actions. In the simulated continuous alphabet task, TD3 gave high performance. However, these results were difficult to achieve consistently, making TD3 impractical for application to the physical robot. 


\section{DISCUSSION}

This work proposes a benchmarking tool aimed at developing the area of reinforcement learning for tactile robotics. Four tasks of varying difficulty are proposed, together with a representative simulated environment to allow for easier exploration of algorithmic adjustments. \textcolor{black}{Evaluation metrics are also provided to allow for a quantitative comparison when using this benchmark with alternative algorithms or alternative sensors.}

We evaluate the performance of several learning-based agents on the proposed benchmark tasks, both in simulation and on the physical robot. We demonstrate that successful learning is possible across all tasks within simulation and across 3 of 4 tasks on the physical robot. Currently, training the physical agent with continuous actions for the full alphabet task has not been achieved within the time allowed by operating constraints in our laboratory. Some example techniques that have not yet been implemented include using prioritised experience replay \cite{schaul2015prioritized} to improve efficiency, and scheduling the ratio of optimisation steps per environment step throughout learning. 

\textcolor{black}{During development, we found some techniques were crucial to achieving successful learning on a physical robot. For example, HER gave sample efficiency improvements up to a factor of 10, along with boosting final performance. This efficiency was particularly evident on the alphabet tasks where rewards are less frequently encountered under random exploration. Training a DQN agent on the simulated discrete alphabet task with HER achieved final performance of $\sim0.98$ within 50 epochs; comparatively, this same task without HER achieved a similar level of performance after $\sim500$ epochs. Thus, HER was required for feasible learning of the tasks on a physical system. When using RL to solve physical robotics problems, designing the tasks to be goal orientated can allow for more general behaviour to be learnt whilst taking advantage of the benefits HER gives. }


\textcolor{black}{
We also found that learnt optimal policies were sensitive to factors other than the algorithm hyper-parameters. For example, when using large action ranges on the continuous arrow tasks, the policies tended towards large movements in the direction of a goal key with low dependence on the current tactile observation. This behaviour arises because a relatively high average return can be found with this method alone, which causes agents to become stuck in a local optimum. Reducing the action ranges to lower values minimised this effect because the sensor was less likely to jump directly to the correct key. Thus, the average return from following this sub-optimal policy was reduced, which ultimately resulted in better policies.}

\textcolor{black}{
Another useful technique was to create a simulation that is partially representative of the final task. Even with recent advancements, DRL is notoriously sensitive and brittle with respect to hyper-parameters, and so a fast, simulated environment helped find parameter regions that allowed for successful and stable training. For example, TD3 was so sensitive that a bad starting seed could cause minimal learning of the task. If attempting to find stable hyper-parameter regions on the physical task, multiple-seeded runs took hours or days of lab operation time. Therefore, where possible, simulating a simplified version of the problem provides valuable information for the physical task. }

\textcolor{black}{
The braille task is designed to be representative of a multitude of tactile robotics tasks in which it may not be practically feasible to create a simulated environment via exhaustive sampling. Therefore, a main aim of this study was to achieve training from scratch in the physical environment. That said, in some circumstances, it would be interesting to explore the benefits of transfer learning from simulation to reality. For a preliminary investigation in simulation, we attempted to capture an important aspect of switching from a simulation to the physical robot, by artificially increasing the step size of the data in the continuous alphabet task from $3\,$mm to $6\,$mm (mimicking that the $3\,$mm intervals are a discrete approximation of the continuous physical environment). We found that learning on the higher density task could be accelerated by transferring policies trained on the lower density task. Whilst this is not entirely aligned to the problem of transfer learning between simulation and reality, these preliminary results demonstrate the potential for large sample-efficiency improvements. However, there are multiple methods of transferring trained policies from simulation to reality, opening up an avenue of future work that uses this platform to examine these sim-to-real approaches.}

Previous research on DRL and tactile data has either used taxel-based sensors \cite{vanHoof2016StableData, Chebotar2016Self-supervisedLearning, wu2019mat, Huang2019LearningLearning}, or has combined optical tactile images with proprioceptive information \cite{Lu2019SurfaceSensor}. To the best of our knowledge, this work presents the first demonstration of successfully training DRL agents in the real world, with observations that exclusively comprise high-dimensional tactile images. Though this work has only presented an evaluation of the TacTip sensor, the proposed experiments could offer valuable comparisons between alternative tactile sensors, particularly in the context of applications using recent DRL techniques. \textcolor{black}{It is possible that alternative tactile sensors could represent the tactile information in a more concise form that allows for more sample-efficient learning, which is the most limiting factor found during this work and an interesting topic for future investigation. To aid with future work, the code used for this project has been released along with designs of the braille keycaps for 3D printing and a guide for recreating experiments and evaluating trained agents (available at \url{https://github.com/ac-93/braille_RL}).}  


\bibliographystyle{IEEEtran}
\bibliography{IEEEabrv, references_updated}

\begin{thebibliography}{10}
\providecommand{\url}[1]{#1}
\csname url@rmstyle\endcsname
\providecommand{\newblock}{\relax}
\providecommand{\bibinfo}[2]{#2}
\providecommand\BIBentrySTDinterwordspacing{\spaceskip=0pt\relax}
\providecommand\BIBentryALTinterwordstretchfactor{4}
\providecommand\BIBentryALTinterwordspacing{\spaceskip=\fontdimen2\font plus
\BIBentryALTinterwordstretchfactor\fontdimen3\font minus
  \fontdimen4\font\relax}
\providecommand\BIBforeignlanguage[2]{{%
\expandafter\ifx\csname l@#1\endcsname\relax
\typeout{** WARNING: IEEEtran.bst: No hyphenation pattern has been}%
\typeout{** loaded for the language `#1'. Using the pattern for}%
\typeout{** the default language instead.}%
\else
\language=\csname l@#1\endcsname
\fi
#2}}

\bibitem{Bellemare2012TheAgents}
M.~G. Bellemare, Y.~Naddaf, J.~Veness, and M.~Bowling, ``The arcade learning
  environment: An evaluation platform for general agents,'' \emph{Journal of
  Artificial Intelligence Research}, vol.~47, pp. 253--279, 2013.

\bibitem{Duan2016BenchmarkingControl}
Y.~Duan, X.~Chen, R.~Houthooft, J.~Schulman, and P.~Abbeel, ``{Benchmarking
  Deep Reinforcement Learning for Continuous Control},'' \emph{International
  Conference on Machine Learning (pp. 1329-1338)}, 4 2016.

\bibitem{Tassa2018DeepMindSuite}
Y.~Tassa, Y.~Doron, A.~Muldal, T.~Erez, Y.~Li, D.~d.~L. Casas, D.~Budden,
  A.~Abdolmaleki, J.~Merel, A.~Lefrancq, T.~Lillicrap, and M.~Riedmiller,
  ``{DeepMind Control Suite},'' \emph{arXiv preprint arXiv:1801.00690}, 1 2018.

\bibitem{Plappert2018Multi-GoalResearch}
M.~Plappert, M.~Andrychowicz, A.~Ray, B.~McGrew, B.~Baker, G.~Powell,
  J.~Schneider, J.~Tobin, M.~Chociej, P.~Welinder, V.~Kumar, and W.~Zaremba,
  ``{Multi-Goal Reinforcement Learning: Challenging Robotics Environments and
  Request for Research},'' \emph{arXiv preprint arXiv:1802.09464}, 2 2018.

\bibitem{Brockman2016OpenAIGym}
G.~Brockman, V.~Cheung, L.~Pettersson, J.~Schneider, J.~Schulman, J.~Tang, and
  W.~Zaremba, ``{OpenAI Gym},'' \emph{arXiv preprint arXiv:1606.01540}, 6 2016.

\bibitem{Ahn2019ROBEL:Robots}
M.~Ahn, H.~Zhu, K.~Hartikainen, H.~Ponte, A.~Gupta, S.~Levine, and V.~Kumar,
  ``{ROBEL:} robotics benchmarks for learning with low-cost robots,'' in
  \emph{3rd Annual Conference on Robot Learning, CoRL,}, ser. Proceedings of
  Machine Learning Research, vol. 100.\hskip 1em plus 0.5em minus 0.4em\relax
  {PMLR}, 2019, pp. 1300--1313.

\bibitem{Kumar2019OffWorldResearch}
A.~Kumar, T.~Buckley, Q.~Wang, A.~Kavelaars, and I.~Kuzovkin, ``{OffWorld Gym:
  open-access physical robotics environment for real-world reinforcement
  learning benchmark and research},'' \emph{arXiv preprint arXiv:1910.08639},
  10 2019.

\bibitem{Yang2019REPLAB:Learning}
B.~Yang, D.~Jayaraman, J.~Zhang, and S.~Levine, ``Replab: A reproducible
  low-cost arm benchmark for robotic learning,'' in \emph{2019 International
  Conference on Robotics and Automation (ICRA)}.\hskip 1em plus 0.5em minus
  0.4em\relax IEEE, 2019, pp. 8691--8697.

\bibitem{Mahmood2018SettingRobot}
A.~R. Mahmood, D.~Korenkevych, B.~J. Komer, and J.~Bergstra, ``{Setting up a
  Reinforcement Learning Task with a Real-World Robot},'' in \emph{2018
  IEEE/RSJ International Conference on Intelligent Robots and Systems (IROS)},
  3 2018, pp. 4635--4640.

\bibitem{Mahmood2018BenchmarkingRobots}
A.~R. Mahmood, D.~Korenkevych, G.~Vasan, W.~Ma, and J.~Bergstra, ``Benchmarking
  reinforcement learning algorithms on real-world robots,'' in \emph{2nd Annual
  Conference on Robot Learning (CoRL),}, ser. Proceedings of Machine Learning
  Research, vol.~87.\hskip 1em plus 0.5em minus 0.4em\relax {PMLR}, 2018, pp.
  561--591.

\bibitem{Mnih2013PlayingLearning}
V.~Mnih, K.~Kavukcuoglu, D.~Silver, A.~Graves, I.~Antonoglou, D.~Wierstra, and
  M.~Riedmiller, ``{Playing Atari with Deep Reinforcement Learning},''
  \emph{arXiv preprint arXiv:1312.5602}, 12 2013.

\bibitem{Fujimoto2018AddressingMethods}
S.~Fujimoto, H.~Hoof, and D.~Meger, ``Addressing function approximation error
  in actor-critic methods,'' in \emph{International Conference on Machine
  Learning}, 2018, pp. 1582--1591.

\bibitem{Haarnoja2018SoftApplications}
T.~Haarnoja, A.~Zhou, K.~Hartikainen, G.~Tucker, S.~Ha, J.~Tan, V.~Kumar,
  H.~Zhu, A.~Gupta, P.~Abbeel, and S.~Levine, ``{Soft Actor-Critic Algorithms
  and Applications},'' \emph{arXiv preprint arXiv:1812.05905}, 12 2018.

\bibitem{James2018Sim-to-RealNetworks}
S.~James, P.~Wohlhart, M.~Kalakrishnan, D.~Kalashnikov, A.~Irpan, J.~Ibarz,
  S.~Levine, R.~Hadsell, and K.~Bousmalis, ``{Sim-to-Real via Sim-to-Sim:
  Data-efficient Robotic Grasping via Randomized-to-Canonical Adaptation
  Networks},'' \emph{Proceedings of the IEEE Computer Society Conference on
  Computer Vision and Pattern Recognition}, vol. 2019-June, pp.
  12\,619--12\,629, 12 2018.

\bibitem{OpenAI2018LearningManipulation}
{OpenAI}, M.~Andrychowicz, B.~Baker, M.~Chociej, R.~Jozefowicz, B.~McGrew,
  J.~Pachocki, A.~Petron, M.~Plappert, G.~Powell, A.~Ray, J.~Schneider,
  S.~Sidor, J.~Tobin, P.~Welinder, L.~Weng, and W.~Zaremba, ``{Learning
  Dexterous In-Hand Manipulation},'' \emph{The International Journal of
  Robotics Research}, vol.~39, no.~1, pp. 3--20, 8 2018.

\bibitem{Rajeswaran2018LearningDemonstrations}
A.~Rajeswaran, V.~Kumar, A.~Gupta, G.~Vezzani, J.~Schulman, E.~Todorov, and
  S.~Levine, ``Learning complex dexterous manipulation with deep reinforcement
  learning and demonstrations,'' in \emph{Robotics: Science and Systems XIV},
  2018.

\bibitem{OpenAI2019SolvingHand}
{OpenAI}, I.~Akkaya, M.~Andrychowicz, M.~Chociej, M.~Litwin, B.~McGrew,
  A.~Petron, A.~Paino, M.~Plappert, G.~Powell, R.~Ribas, J.~Schneider,
  N.~Tezak, J.~Tworek, P.~Welinder, L.~Weng, Q.~Yuan, W.~Zaremba, and L.~Zhang,
  ``{Solving Rubik's Cube with a Robot Hand},'' \emph{arXiv preprint
  arXiv:1910.07113}, 10 2019.

\bibitem{vanHoof2016StableData}
H.~van Hoof, N.~Chen, M.~Karl, P.~van~der Smagt, and J.~Peters, ``{Stable
  reinforcement learning with autoencoders for tactile and visual data},'' in
  \emph{2016 IEEE/RSJ International Conference on Intelligent Robots and
  Systems (IROS)}, 10 2016, pp. 3928--3934.

\bibitem{Chebotar2016Self-supervisedLearning}
Y.~Chebotar, K.~Hausman, Z.~Su, G.~S. Sukhatme, and S.~Schaal,
  ``{Self-supervised regrasping using spatio-temporal tactile features and
  reinforcement learning},'' in \emph{IEEE International Conference on
  Intelligent Robots and Systems}, vol. 2016-November, 11 2016, pp. 1960--1966.

\bibitem{Huang2019LearningLearning}
S.~H. Huang, M.~Zambelli, J.~Kay, M.~F. Martins, Y.~Tassa, P.~M. Pilarski, and
  R.~Hadsell, ``{Learning Gentle Object Manipulation with Curiosity-Driven Deep
  Reinforcement Learning},'' \emph{arXiv preprint arXiv:1903.08542}, 3 2019.

\bibitem{Lu2019SurfaceSensor}
C.~Lu, J.~Wang, and S.~Luo, ``{Surface Following using Deep Reinforcement
  Learning and a GelSightTactile Sensor},'' \emph{arXiv preprint
  arXiv:1912.00745}, 12 2019.

\bibitem{Lepora2018FromSensor}
N.~F. Lepora, A.~Church, C.~De~Kerckhove, R.~Hadsell, and J.~Lloyd, ``{From
  pixels to percepts: Highly robust edge perception and contour following using
  deep learning and an optical biomimetic tactile sensor},'' \emph{IEEE
  Robotics and Automation Letters}, 12 2018.

\bibitem{Yuan2017Shape-independentSensor}
W.~Yuan, C.~Zhu, A.~Owens, M.~A. Srinivasan, and E.~H. Adelson,
  ``{Shape-independent Hardness Estimation Using Deep Learning and a GelSight
  Tactile Sensor},'' \emph{Sensors}, vol.~17, no.~12, p. 2762, 4 2017.

\bibitem{Calandra2017TheOutcomes}
R.~Calandra, A.~Owens, M.~Upadhyaya, W.~Yuan, J.~Lin, E.~H. Adelson, and
  S.~Levine, ``The feeling of success: Does touch sensing help predict grasp
  outcomes?'' in \emph{1st Annual Conference on Robot Learning (CoRL)}, ser.
  Proceedings of Machine Learning Research, vol.~78.\hskip 1em plus 0.5em minus
  0.4em\relax PMLR, 13--15 Nov 2017, pp. 314--323.

\bibitem{Calandra2018MoreTouch}
R.~Calandra, A.~Owens, D.~Jayaraman, J.~Lin, W.~Yuan, J.~Malik, E.~H. Adelson,
  and S.~Levine, ``{More Than a Feeling: Learning to Grasp and Regrasp using
  Vision and Touch},'' \emph{IEEE Robotics and Automation Letters}, vol.~3,
  no.~4, pp. 3300--3307, 5 2018.

\bibitem{Hogan2018TactileTransformations}
F.~R. Hogan, M.~Bauza, O.~Canal, E.~Donlon, and A.~Rodriguez, ``{Tactile
  Regrasp: Grasp Adjustments via Simulated Tactile Transformations},'' in
  \emph{2018 IEEE/RSJ International Conference on Intelligent Robots and
  Systems (IROS)}, 3 2018, pp. 2963--2970.

\bibitem{Lepora2020IEEERAM}
N.~F. Lepora and J.~Lloyd, ``Optimal deep learning for robot touch: Training
  accurate pose models of 3d surfaces and edges,'' \emph{{IEEE} Robotics Autom.
  Mag.}, vol.~27, no.~2, pp. 66--77, 2020.

\bibitem{Ward-Cherrier2018TheMorphologies}
B.~Ward-Cherrier, N.~Pestell, L.~Cramphorn, B.~Winstone, M.~E. Giannaccini,
  J.~Rossiter, and N.~F. Lepora, ``{The TacTip Family: Soft Optical Tactile
  Sensors with 3D-Printed Biomimetic Morphologies},'' \emph{Soft Robotics},
  vol.~5, no.~2, pp. 216--227, 4 2018.

\bibitem{Andrychowicz2017HindsightReplay}
M.~Andrychowicz, D.~Crow, A.~Ray, J.~Schneider, R.~Fong, P.~Welinder,
  B.~McGrew, J.~Tobin, P.~Abbeel, and W.~Zaremba, ``Hindsight experience
  replay,'' in \emph{Annual Conference on Neural Information Processing
  Systems}, 2017, pp. 5048--5058.

\bibitem{Mnih2015Human-levelLearning}
V.~Mnih, K.~Kavukcuoglu, D.~Silver, A.~A. Rusu, J.~Veness, M.~G. Bellemare,
  A.~Graves, M.~Riedmiller, A.~K. Fidjeland, G.~Ostrovski, S.~Petersen,
  C.~Beattie, A.~Sadik, I.~Antonoglou, H.~King, D.~Kumaran, D.~Wierstra,
  S.~Legg, and D.~Hassabis, ``{Human-level control through deep reinforcement
  learning},'' \emph{Nature}, vol. 518, no. 7540, pp. 529--533, 2 2015.

\bibitem{Schulman2015TrustOptimization}
J.~Schulman, S.~Levine, P.~Moritz, M.~I. Jordan, and P.~Abbeel, ``{Trust Region
  Policy Optimization},'' \emph{International conference on machine learning
  (pp. 1889-1897)}, 2 2015.

\bibitem{Schulman2017ProximalAlgorithms}
J.~Schulman, F.~Wolski, P.~Dhariwal, A.~Radford, and O.~Klimov, ``{Proximal
  Policy Optimization Algorithms},'' \emph{arXiv preprint arXiv:1707.06347}, 7
  2017.

\bibitem{Lillicrap2015ContinuousLearning}
T.~P. Lillicrap, J.~J. Hunt, A.~Pritzel, N.~Heess, T.~Erez, Y.~Tassa,
  D.~Silver, and D.~Wierstra, ``Continuous control with deep reinforcement
  learning,'' in \emph{4th International Conference on Learning
  Representations, {ICLR}}, 2016.

\bibitem{Haarnoja2018SoftActor}
T.~Haarnoja, A.~Zhou, P.~Abbeel, and S.~Levine, ``Soft actor-critic: Off-policy
  maximum entropy deep reinforcement learning with a stochastic actor,'' in
  \emph{International Conference on Machine Learning}, 2018, pp. 1861--1870.

\bibitem{Abdolmaleki2018MaximumOptimisation}
A.~Abdolmaleki, J.~T. Springenberg, Y.~Tassa, R.~Munos, N.~Heess, and M.~A.
  Riedmiller, ``Maximum a posteriori policy optimisation,'' in \emph{6th
  International Conference on Learning Representations, {ICLR}}.\hskip 1em plus
  0.5em minus 0.4em\relax OpenReview.net, 2018.

\bibitem{Hessel2017Rainbow:Learning}
M.~Hessel, J.~Modayil, H.~van Hasselt, T.~Schaul, G.~Ostrovski, W.~Dabney,
  D.~Horgan, B.~Piot, M.~G. Azar, and D.~Silver, ``Rainbow: Combining
  improvements in deep reinforcement learning,'' in \emph{Proceedings of the
  Thirty-Second {AAAI} Conference on Artificial Intelligence}.\hskip 1em plus
  0.5em minus 0.4em\relax {AAAI} Press, 2018, pp. 3215--3222.

\bibitem{Hasselt2016DeepQ-Learning}
H.~van Hasselt, A.~Guez, and D.~Silver, ``Deep reinforcement learning with
  double q-learning,'' in \emph{Proceedings of the Thirtieth {AAAI} Conference
  on Artificial Intelligence}.\hskip 1em plus 0.5em minus 0.4em\relax {AAAI}
  Press, 2016, pp. 2094--2100.

\bibitem{Wang2015DuelingLearning}
Z.~Wang, T.~Schaul, M.~Hessel, H.~van Hasselt, M.~Lanctot, and N.~de~Freitas,
  ``Dueling network architectures for deep reinforcement learning,'' in
  \emph{Proceedings of the 33nd International Conference on Machine Learning,
  {ICML}}, ser. {JMLR} Workshop and Conference Proceedings, vol.~48.\hskip 1em
  plus 0.5em minus 0.4em\relax JMLR.org, 2016, pp. 1995--2003.

\bibitem{Popov2017Data-efficientManipulation}
I.~Popov, N.~Heess, T.~Lillicrap, R.~Hafner, G.~Barth-Maron, M.~Vecerik,
  T.~Lampe, Y.~Tassa, T.~Erez, and M.~Riedmiller, ``{Data-efficient Deep
  Reinforcement Learning for Dexterous Manipulation},'' \emph{arXiv preprint
  arXiv:1704.03073}, 4 2017.

\bibitem{schaul2015prioritized}
T.~Schaul, J.~Quan, I.~Antonoglou, and D.~Silver, ``Prioritized experience
  replay,'' in \emph{4th International Conference on Learning Representations,
  {ICLR}}, 2016.

\bibitem{wu2019mat}
B.~Wu, I.~Akinola, J.~Varley, and P.~K. Allen, ``{MAT:} multi-fingered adaptive
  tactile grasping via deep reinforcement learning,'' in \emph{3rd Annual
  Conference on Robot Learning (CoRL)}, ser. Proceedings of Machine Learning
  Research, vol. 100.\hskip 1em plus 0.5em minus 0.4em\relax {PMLR}, 2019, pp.
  142--161.

\end{thebibliography}

\end{document}